# AN FFT-BASED SYNCHRONIZATION APPROACH TO RECOGNIZE HUMAN BEHAVIORS USING STN-LFP SIGNAL

Hosein M. Golshan[a], Adam O. Hebb[b], Sara J. Hanrahan[b],
Joshua Nedrud[b], Mohammad H. Mahoor[a], *Senior Member, IEEE*

[a]CV Lab, ECE Department, University of Denver, CO, USA
[b]Colorado Neurological Institute, Englewood, CO, USA

## ABSTRACT

Classification of human behavior is key to developing closed-loop Deep Brain Stimulation (DBS) systems, which may be able to decrease the power consumption and side effects of the existing systems. Recent studies have shown that the Local Field Potential (LFP) signals from both Subthalamic Nuclei (STN) of the brain can be used to recognize human behavior. Since the DBS leads implanted in each STN can collect three bipolar signals, the selection of a suitable pair of LFPs that achieves optimal recognition performance is still an open problem to address. Considering the presence of synchronized aggregate activity in the basal ganglia, this paper presents an FFT-based synchronization approach to automatically select a relevant pair of LFPs and use the pair together with an SVM-based MKL classifier for behavior recognition purposes. Our experiments on five subjects show the superiority of the proposed approach compared to other methods used for behavior classification.

## 1. INTRODUCTION

Intracortical microelectrode recordings from the basal ganglia provide access to a variety of neural signals such as single/multi-unit activity and Local Field Potentials (LFPs) [1]. In particular, the opportunity arises in patients with Parkinson's Disease (PD is a neurodegenerative disorder pertaining to the central nervous system) undergoing Deep Brain Stimulation (DBS) surgery, which requires implantation of the DBS leads in the Subthalamic Nuclei (STN) for therapeutic stimulation [2]. Signals acquired from the DBS leads can be used for characterization of the human activities within cortical regions and sub-cortical nuclei [3].

Decoding human behavior using different brain signals has gained increasing attention in recent years. A number of Electroencephalography (EEG)-based methods have been developed in which patterns of the EEG signals in different mental states can be recognized for information transmission by feature extraction and classification methods [4-7]. Many studies have focused on the real time detection of behavior using EEG and Electrocorticography (ECoG) data such as P300 detection for spelling [8,9], brain-switch based on motor imagery [10,11], and self-regulation of rhythm [5].

Developing a closed-loop DBS system capable of customizing the stimulation parameters has recently been presented in the literatures [12-14]. Note that, in spite of its remarkable performance in providing relief of PD's motor symptoms, e.g., tremor and rigidity, DBS may generate some side effects such as cognitive and balance disruptions mainly due to the existing open-loop systems [3]. In an attempt to design a closed-loop DBS system, a number of studies have been dedicated to the human behavior recognition using LFP signals. Loukas and Brown [13] proposed an algorithm to predict self-paced hand-movements from the oscillatory nature of the STN-LFPs. Santaniello *et al.,* [14] presented a closed-loop DBS system capable of adjusting the stimulation amplitude using the LFP feedback from Ventral Intermediate Nucleus (VIM) of the thalamus. The time-frequency analysis of LFPs has been considered to classify different human behavioral activities [15-17]. Several classification methods based on Hidden Markov Model (HMM) and Deep Neural Network have been suggested in [18-20].

In this paper, we present a human behavior recognition approach using the time-frequency analysis (spectrogram) of STN-LFP signals. Note that, through the DBS surgery two DBS leads, each of them including four contacts, are implanted in the STN regions. Thus, for each trial 6 bipolar channels (i.e., three channels from each STN) are defined to collect the corresponding LFP signals. Contrary to other related works [15-17] that select a pair of LFP signals from the left and right bipolar channels regardless of their mutual interplay, here, we apply an FFT-based synchronization method [21,22] as our guideline to choose a relevant pair of LFPs for each subject under study. This in return can lead to the bipolar channels with the most informative LFP signals that are in the optimal location in the STN.

There is evidence that LFPs recorded from the basal ganglia reflect synchronized aggregate activity [13]. This has also been supported by studies in Parkinsonian patients [23-25]. Inter-hemispheric synchronization occurs in several frequency bands partly dictated by the level of dopaminergic stimulation [13]. In our experiments, we observed that using the synchronization approach for data arrangement leads to higher recognition accuracies with any classifiers in use, including the widely used support vector machines equipped with Multiple Kernel Learning (MKL).

The rest of this paper is organized as follows: Section II presents the data recording procedure in details. Section III elaborates the proposed approach. Section IV provides the comparisons and quantitative assessments. Conclusions and some remarks are given in Section V.

## 2. RECORDING DESIGN

### 2.1. Subjects

Five subjects undergoing DBS surgery as standard of care for treatment of idiopathic PD were enrolled in this study. All subjects provided informed consent for participation in this research in a manner approved by the HealthOne Institutional Review Board. LFP signals were collected from all four contacts of the bilaterally implanted DBS leads (*Medtronic 3389*, *Minneapolis*, *MN*, *USA*). Note that, all patients were in the off medication state, and we did not proceed with recording until patients were fully awake during surgery [2]. On average, each data acquisition session lasted about 30 minutes. The collected data was amplified, digitized (4.8 kHz), band passed filtered (1-100Hz), and combined with event markers and subject responses. A linked-mastoid common reference was used for recordings. Finally, the LFP signals were bipolar re-referenced (0-1, 1-2, 2-3) to generate three bipolar signals for each STN.

### 2.2. Behavioral Studies

Behavior included button press, arm movement, speech, and mouth movement. For each behavior, a block of several cued repetitions was performed. "Button press" consisted of pressing a button using either the left or right thumb. "Speech" included repeating object names displayed on the screen. "Arm movement" required the patients to raise their arm to reach a target appearing on the screen using either the left or right hand. Finally, as a comparison to the "Speech" trial, "Mouth movement" was simply composed of moving the mouth without speech.

## 3. METHODS

This section describes the proposed method in details. Our method utilizes an FFT-based synchronization approach together with the SVM-based MKL classifier for human behavior recognition purposes. Fig. 1 shows a block diagram of the proposed classification approach.

### 3.1. FFT-based Synchronization

Here, we use an FFT-based approach to find the most synchronous pair of LFP signals in each case [21,22], providing more reliable dataset for training the employed classifiers. Note that, a signal can have many phase values associated with each Fourier components. The FFT-based synchronization considers the phase values of each frequency component separately, leading to a more minute measure of phase synchronization based on a finer resolution compared to the statistical correlation-based measures [21]. Moreover, it is independent of the amplitude of signal and takes no longer than the FFT algorithm.

Assuming two continuous LFP signals $x_i(t)$ and $x_j(t)$ acquired from the left and right STNs, the FFT synchronization measure is calculated by extracting all frequency components of these signals. Considering the Fourier coefficients $a_{in}$, $a_{jn}$ and $b_{in}$, $b_{jn}$ calculated respectively for the $n^{th}$ frequency component of signals $x_i(t)$ and $x_j(t)$, the corresponding phase values are given by:

$$\theta_{in} = \tan^{-1}\left(\frac{a_{in}}{b_{in}}\right), \quad \theta_{jn} = \tan^{-1}\left(\frac{a_{jn}}{b_{jn}}\right). \quad (1)$$

The main idea behind the FFT synchronization method is that, the phase lag of two synchronous signals should be almost uniform across all harmonics [21]. As a consequence, for two approximately phase synchronous signals the corresponding phase components $\theta_{in}$ and $\theta_{jn}$ are almost equal. So, for the $n^{th}$ frequency component the phase lag (*PL*) value can easily be calculated by the corresponding Fourier coefficients:

$$PL(n) = \left|\theta_{in} - \theta_{jn}\right| \approx 0 \Rightarrow PL(n) = \left|\frac{a_{in}b_{jn} - b_{in}a_{jn}}{a_{in}a_{jn} + b_{in}b_{jn}}\right| \approx 0. \quad (2)$$

To obtain an accurate estimate of the phase synchronization between two signals, the phase lag values for all harmonics should be taken into account. This implies both the mean and standard deviation of the calculated phase lag values to be a small quantity for two nearly synchronous signals, resulting in a small phase difference for each frequency component. As stated in [21], the FFT synchronization for two signals $x_i(t)$ and $x_j(t)$ is given by:

$$sync(x_i(t), x_j(t)) = \frac{1}{1+mean(E(n))+std(E(n))}. \quad (3\text{-a})$$

$$E(n) = |PL(n) - PL(n+1)| =$$
$$\left| \frac{a_{in}b_{jn} - b_{in}a_{jn}}{a_{in}a_{jn} + b_{in}b_{jn}} - \frac{a_{in+1}b_{jn+1} - b_{in+1}a_{jn+1}}{a_{in+1}a_{jn+1} + b_{in+1}b_{jn+1}} \right|. \quad (3\text{-b})$$

where, $mean(\cdot)$ and $std(\cdot)$ are respectively the average and standard deviation of the quantity $E(n)$ calculated across all the frequency components. Eq. (3-a) guarantees that the synchronization values are normalized in the range of [0, 1], so the more phase synchronous two signals are the closer to 1 is the value of $sync(\cdot)$.

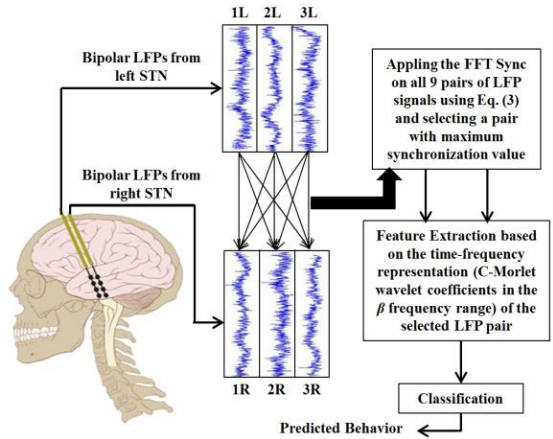

Fig.1. Block diagram of the proposed behavior classification scheme using the FFT-based synchronization approach. Each arrow between the left and right LFP signals shows the corresponding pair selected for the synchronization step.

To apply the above-described synchronization method on the collected LFP signals, Eq. (3-a, 3-b) are calculated for all 9 possible LFP pairs before any analysis takes place (see Fig. 1). As mentioned earlier, one of the main advantages of the FFT-based synchronization is its low computational complexity, which is equal to that of the FFT algorithm [21]. As a result, this approach can automatically lead us to the near optimal LFP pairs for each subject without imposing any further computational burden.

### 3.2. Classification Scheme

The main contribution of this paper is to provide a method to properly select a pair of LFP signals for human behavior classification irrespective of the employed classifier and subject under study. However, in our experiments, we focus on a recently proposed SVM-based MKL classifier, as it was shown to obtain promising results for behavior classification using STN-LFP signals [17].

The SVM-based MKL classifier aims to optimally combine matrices calculated based on multiple features with multiple kernels in SVM [26-30]. In other words, it learns both the decision boundaries between different classes and kernel combination weights in a single optimization problem, improving the discriminant power of the SVM [27,28]. Here, we utilize an $lp$-norm realization of the MKL formulation which proved to be more flexible in selecting different kernel combinations. It is given by:

$$\min_{\omega, \omega_0, \xi} J(\omega, \omega_0, \xi) = \frac{1}{2} \|\omega\|_{2,p}^2 + C\sum_{i=1}^{N}\xi_i. \quad (4)$$

$$\text{s.t. } y_i\left(\sum_{m=1}^{M}\omega_m^T\varphi_m(\mathbf{x}_i)+\omega_0\right) \geq 1-\xi_i, \quad \xi_i \geq 0, p \geq 1, i=1,\ldots,N$$

where, $\varphi_m(\cdot)$ maps the feature vector $\mathbf{x}_i$ to another space based on which the kernel function $\mathbf{K}(\cdot,\cdot) = \langle\varphi_m(\cdot), \varphi_m(\cdot)\rangle$ is defined. $\{\omega_m\}$s are the parameters of the decision hyper-planes. $M$ and $N$ are the number of kernels and training samples respectively. $C$ is the penalty parameter and $\xi_i$ is the vector of slack variables. The parameter $p$ in Eq. (4) is to regularize over kernel combination coefficients, which considers both sparse and non-sparse kernel combinations within MKL. Note that, this convex optimization problem is solved using its dual form (the readers are referred to [28,29] for more details on the definition of parameters as well as the dual form equations). Consequently, the label $y_z$ for each test sample $z \in R^d$ can be calculated by:

$$y_z = \text{sgn}\left(\sum_{i=1}^{N}\sum_{m=1}^{M}\alpha_i y_i \mathbf{d}_m \mathbf{K}_m(\mathbf{x}_i, z) + \omega_0\right). \quad (5)$$

where, $\alpha$ is the vector of Lagrangian dual multipliers and $\mathbf{d}$ is the kernel combination vector that controls the weight of $(\|\omega\|^2)$ in the objective function of Eq. (4).

Table I. Comparison of the average classification accuracy for all subjects. In all cases, the FFT synchronization approach together with the employed classifier leads to the best results (second row).

| | SVM-linear | SVM-Polynomial | SVM-RBF | MKL |
|---|---|---|---|---|
| Without Sync | 54.53 | 35.18 | 28.78 | 57.17 |
| FFT Sync | 57.53 | 36.67 | 30.39 | 61.00 |

## 4. EXPERIMENTS AND RESULTS

To assess the effect of the given FFT-based synchronization approach on the classification of different human behavior, we used the raw LFP signals collected from 5 subjects undergoing DBS surgery (see Section II for more details). To calculate the synchronization value for all 9 possible LFP pairs, Eq. (3-a, 3-b) were applied on the acquired LFP signals (i.e., the FFT analysis was performed on the entire LFP signal, no sliding window was used with our calculations); the maximum synchronization value gives us the desired pair of LFP signals for any further post-processing steps as well as classification.

It has been shown [2,3] that $\beta$ frequency range (13-30Hz) of LFP signals is an appropriate feature to discriminate different human behavior in the time-frequency domain. Therefore, we used the complex Morlet wavelet, which proved to be a suitable method for biomedical signal processing, to calculate the spectrogram of the raw LFP signals [2,15,17]. For each trial the wavelet coefficients in the $\beta$ frequency range, calculated inside the [-1, 1] seconds interval around the onset, was used as the feature vector. Afterwards, we down-sampled feature vectors by a factor of 100 and applied Principal Component Analysis (PCA) on the down-sampled data to minimize the computational cost (in each case, 95% of the eigenvalues corresponding to the maximum variance direction was kept).

The effect of the FFT-based synchronization on the human behavior classification was evaluated using two recently proposed SVM [15] and MKL [17] approaches. We also studied the effect of various kernel functions on the classification performance, including linear $\mathbf{K}(\mathbf{x}, \mathbf{y}) = \mathbf{x}^T\mathbf{y}+c$, polynomial $\mathbf{K}(\mathbf{x}, \mathbf{y}) = (\mathbf{x}^T\mathbf{y}+c)^d$, and RBF $\mathbf{K}(\mathbf{x}, \mathbf{y}) = exp(\gamma\|\mathbf{x}-\mathbf{y}\|^2)$ kernels [27]. Note that, $\mathbf{x}$ and $\mathbf{y}$ are two feature vectors, and $\gamma$, $c$, and $d$ are optional constants. In terms of the $l_p$-norm MKL, we set the parameters $C=100$ and $p=1.5$ to achieve the best performance. A leave-one-out cross validation was implemented in all experiments [29].

Table I provides the average classification accuracies for all 5 subjects performing, "button press", "speech", "arm movement", "mouth movement", and "rest segment". The "rest segment" contains those segments of the LFP signals where the patient is not doing any activity. We add the "rest segment" to our experiments to train the classifiers to recognize other behavior as well. As shown in Table I, regardless of the classifier in use, in all cases the best results are obtained using the pairs of LFP signals given by the synchronization approach, likely due to selecting bipolar channels with more informative signals in the sensorimotor area of the STN. For example, MKL achieves 61% classification accuracy using the synchronization approach while the average accuracy of this method is 57.17%.

Fig. 2 evaluates the efficiency of the synchronization approach against each of the 9 possible LFP pairs separately. Each bar in the figure represents the average classification accuracy for all subjects. As seen, the LFP pair selected by the FFT synchronization method leads to the best result in comparison with other possible pairs. In particular, the average classification accuracy given by the LFP pair 3L-2R (60.22%) is comparable to that of the FFT synchronization approach (61%). However, while the synchronization method can automatically select the optimal LFP pair for each subject without imposing a considerable computational cost (i.e., the computational time is no longer than the FFT algorithm), one should repeat the time-consuming training and validation phases for all possible LFP pairs to get the optimal pair in each case. Table II shows the average confusion matrix of all subjects using the FFT-based synchronization approach and the SVM-based MKL classifier, which summarizes the identification results. As seen, the highest recognition accuracy is for the "Arm movement" behavior (80%) while "Button Press" and "Mouth Movement" are the most difficult cases (53%).

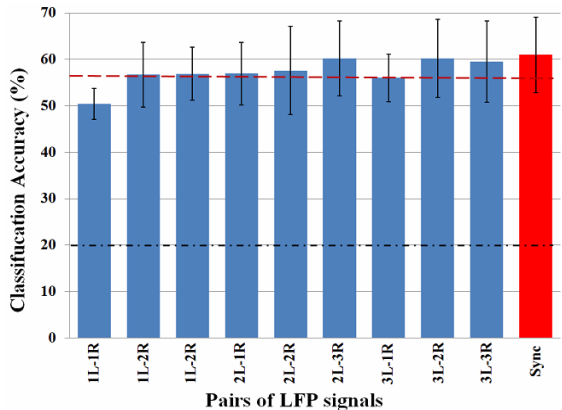

Fig. 2. Average Classification accuracy of the MKL approach for all LFP pairs. The red bar shows the result using the FFT synchronization. Red dash-line shows the average of all 9 blue bars, and the black dash-dot line represents the chance rate.

Table II. Average confusion matrix (normalized in [0,100]) for all subjects and the MKL approach. Row and Column directions respectively show the ground-truth and predicted behavior.

|    | BP | S  | RS | AM | MM |                    |
|----|----|----|----|----|----|--------------------|
| BP | **53** | 12 | 11 | 12 | 12 | BP: Button Press   |
| S  | 9  | **60** | 10 | 2  | 19 | S: Speech          |
| RS | 12 | 11 | **57** | 7  | 13 | RS: Rest Segment   |
| AM | 10 | 1  | 3  | **80** | 6  | AM: Arm Movement   |
| MM | 15 | 19 | 11 | 2  | **53** | MM: Mouth Movement |

## 5. CONCLUSION AND DISCUSSION

In this paper, an FFT-based synchronization approach was presented to automatically select a pair of the bipolar LFP signals from 9 available pairs. The selected pair was then used as the signal for human behavior recognition, which is of great importance for designing the next generation of closed-loop DBS systems. Note that, each DBS lead implanted in the left or right subthalamic nuclei of the brain can collect 3 bipolar LFP signals. Inherently, some of the acquired signals are less informative than others, likely due to their location in the sensorimotor area of the STN. So, they cannot be appropriate candidates for post-processing purposes. In our proposed method, however, the pairs of LFPs with highest synchronization values were considered for human behavior classification, leading to better classification accuracy.

We evaluated the effect of the synchronization approach on the behavior recognition using single kernel SVM as well as SVM-based MKL classifiers. The experiments were carried out on the LFP signals acquired from 5 subjects undergoing DBS surgery. The classification performance for different human behavior including button press, arm movement, speech, mouth movement, and rest state was studied in this work. Regardless of the employed classifier, the synchronization approach improved the behavior classification accuracy in all cases mainly due to the more reliable dataset provided for training phase.

Evaluating the connectivity between different parts of the brain using synchronization measures can be an interesting topic for the future research. Expanding this method to using other kinds of brain signals (e.g., ECoG from pre-frontal cortex) with the behavior classification task can potentially enhance the classification performance.

## ACKNOWLEDGMENT

The authors gratefully acknowledge support from the Colorado Neurological Institute (CNI) and Swedish Medical Center, Englewood, Colorado. This research is supported by the Knoebel Institute for Healthy Aging at University of Denver.